**Optimizing for the decision not the prediction: an exploration of Smooth Net Benefit as a training objective.**

**Koen K.M.F. Gorgels[1,2], Lasai Barreñada[3], Maarten van Smeden[1], Ben Van Calster[1,3], Ewout W. Steyerberg[1], Wouter A.C. van Amsterdam[1]**

[1] Department of Data Science and Biostatistics, Julius Center for Health Sciences and Primary Care, University Medical Center Utrecht, Utrecht, the Netherlands

[2] Coöperatie VGZ, Arnhem, the Netherlands

[3] Department of Development and Regeneration, KU Leuven, Leuven, Belgium

## Abstract

### Objective

Prediction models are commonly trained using objectives such as Bernoulli negative log-likelihood (NLL), although downstream clinical decisions may depend on specific risk thresholds. We introduce Smooth Net Benefit (σNB), a differentiable approximation of Net Benefit designed to align model training with threshold-specific clinical utility.

### Materials and Methods

We evaluated σNB as a training objective for logistic regression, generalized additive models (GAMs), and XGBoost with three Hessian implementations. Experiments used the Framingham cardiovascular risk dataset and 44 TabZilla datasets comprising 72 dataset–threshold combinations.

### Results

σNB training did not consistently improve Net Benefit in Framingham. Across the TabZilla benchmark, mean standardized Net Benefit for logistic regression increased from 0.5669 with NLL to 0.5765 with sNB (mean difference 0.0096, 95% CI −0.0001 to 0.0193). For GAMs, mean standardized Net Benefit decreased from 0.5921 to 0.5625 (mean difference −0.0296, 95% CI −0.0721 to 0.0129). For XGBoost, NLL achieved 0.6745 compared with 0.6723–0.6735 across σNB implementations. In logistic regression, sNB gains were positively associated with the performance advantage of XGBoost over NLL-trained logistic regression.

### Discussion

The effect of σNB was context dependent, with modest gains concentrated in logistic regression and little benefit for more flexible model classes. This suggests that decision-focused optimization may be most useful when limited model flexibility leaves greater scope for improvement.

### Conclusion

Our results do not support σNB as a general replacement for NLL training, but support further investigation of decision-focused objectives in settings where conventional likelihood-based training may not adequately capture decision-relevant structure.

# 1 Background and significance

Prediction models are used in medical decision-making, for example by recommending treatment, further diagnostic testing, or switching interventions when an individual's estimated risk exceeds a predefined threshold(1–3). Such thresholds reflect a trade-off between correctly identifying patients who may benefit from additional treatment or testing and avoiding unnecessary interventions in patients at low risk of adverse outcomes. There is increasing recognition that prediction models should be evaluated not only by their predictive performance, but also by their impact on downstream decisions (4–7).

A widely used framework for evaluating prediction-guided decisions is Net Benefit (NB), originally developed as part of decision curve analysis to quantify the clinical utility of prediction models (1–3). A central concept in decision curve analysis is the specification of a relevant risk threshold $t$, or range of thresholds, at which decisions are made (8). Predictions above the threshold trigger a change in policy, such as treatment or further diagnostic testing, while the threshold itself reflects the relative value assigned to beneficial and harmful decisions. In the standard binary-outcome formulation, NB is the proportion of true-positive decisions minus the proportion of false-positive decisions weighted by the threshold odds. Consequently, model behavior around thresholds relevant to the intended clinical use is particularly important for downstream decision-making.

Despite its relevance to decision-making, NB is used as an evaluation metric, while model training typically relies on objectives such as the (Bernoulli) Negative Log Likelihood (NLL) that optimize global predictive performance. This may leave clinically relevant threshold regions under-prioritized during training (3,9). Related work on decision-aware calibration has shown that model behavior can be selectively improved near plausible decision boundaries rather than uniformly across the full probability space (10). This suggests that decision performance may potentially be improved through training procedures that place greater emphasis on these regions during model development.

Optimal threshold-based decisions require less information about the conditional outcome probability than optimal probabilistic prediction: at a fixed decision threshold, the relevant question is whether an individual's risk lies above or below that threshold, whereas NLL rewards probability estimation across the full probability range. We therefore hypothesized that directly targeting Net Benefit could improve decision performance relative to NLL training. To investigate this, we introduce Smooth Net Benefit (σNB), a differentiable approximation of NB that replaces the hard decision threshold with a smooth approximation. Because this objective is non-convex and provides limited gradient information when predictions are far from the decision threshold, σNB training is warm-started from an NLL-trained model and uses temperature annealing to progressively reduce smoothing. We evaluate σNB for logistic regression, generalized additive models (GAMs), and XGBoost using the Framingham cardiovascular dataset and a large benchmark of tabular datasets, with an additional small-sample analysis.

# 2 Materials and methods

## 2.1 Smooth Net Benefit

We consider prediction models $f$ for a binary outcome $Y$ that given features $X$ aim to estimate the probability $0 \leq P(Y=1|X=x) \leq 1$. At a threshold $0 \leq t \leq 1$, define true positives for a sample of observations $(x_i, y_i)_{i=1}^{n}$ as $TP_t = \sum_i y_i I_{f(x_i)>t}$ and false positives as $FP_t = \sum_i (1-y_i) I_{f(x_i)>t}$, then the Net Benefit at decision threshold $t$ is defined as (1):

$$NB_t(f) = \frac{TP_t(f)}{N} - \frac{FP_t(f)}{N} * w_t$$

where $w_t = \frac{t}{1-t}$ is the weight assigned to false positives at threshold $t$, which is equivalent to:

$$\mathrm{NB}_t(f) = \frac{1}{N}\sum_{i=1}^{N} I_{f(X_i)>\mathrm{t}}\,(y_i - (1-y_i) * w_t) \qquad (1)$$

Note that models for which $f(x) \leq t$ for all patients have 0 net benefit by definition.

For a model with parameters $\theta$, net benefit is a piecewise constant function of $\theta$, with zero gradient throughout most of the parameter space and discontinuities whenever predictions cross the decision threshold. This makes net benefit difficult to optimize directly. Instead, we introduce a smooth approximation using sigmoid smoothing, Smooth Net Benefit (σNB):

$$\sigma\mathrm{NB}_{t,k}(f) = \frac{1}{N}\sum_{i=1}^{N} \sigma\,(k * (f(X_i) - t))(y_i - (1-y_i) * w_t) \qquad (2)$$

where $\sigma$ is the sigmoid (logistic) function $\sigma(x) = 1/(1+e^{-x})$ and $k$ is an inverse-temperature parameter controlling the steepness of the approximation. By increasing $k$, σNB becomes a closer approximation of NB, but has non-zero gradients everywhere, making it amenable to optimization with gradient-based methods. See Figure 1 for a visualization of σNB as a function of predicted risk.

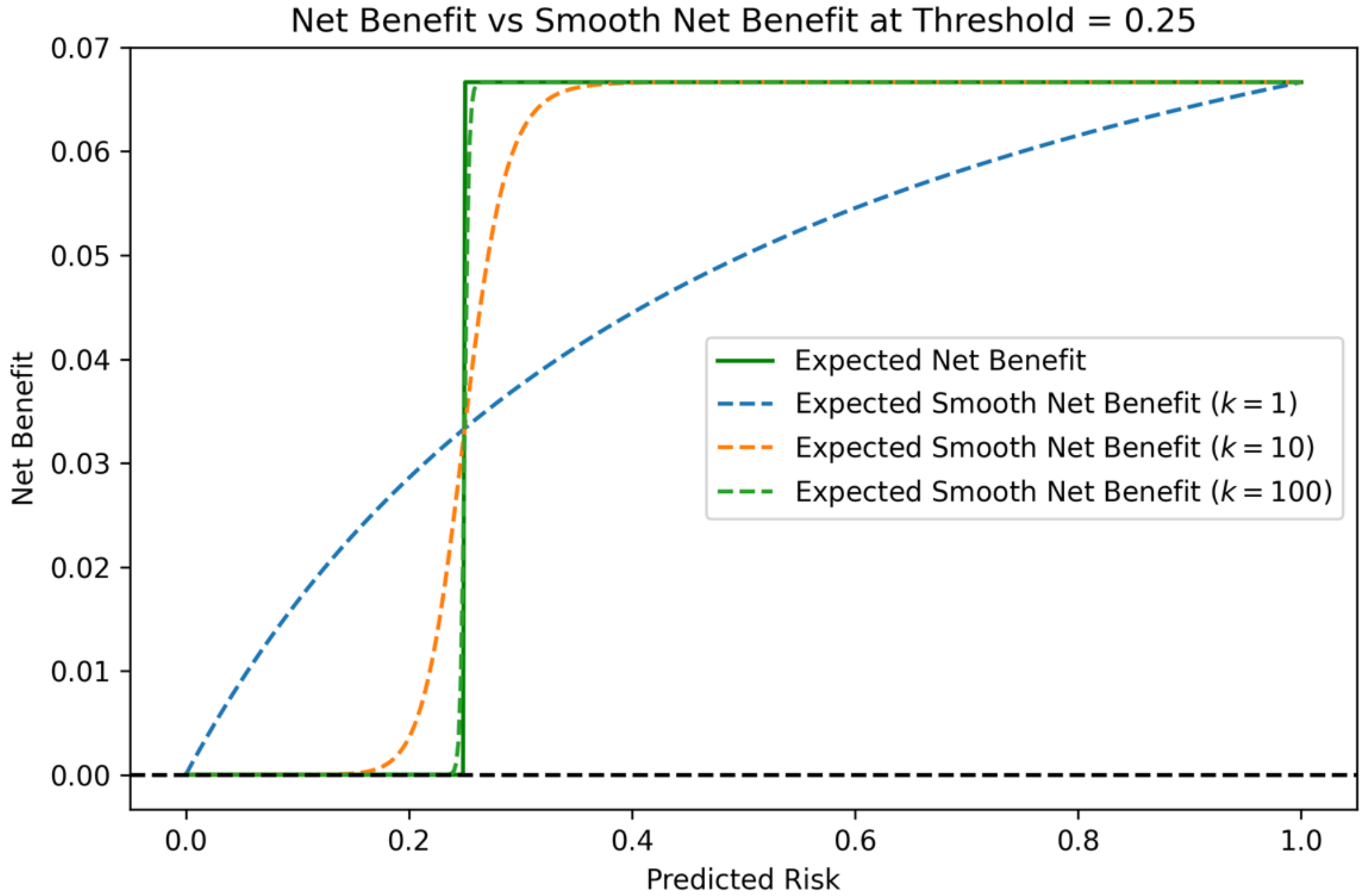

*Figure 1. Net Benefit versus Smooth Net Benefit in a simple setting where the only parameter is the average predicted risk for the positive class (displayed on the horizontal axis), and the expected net benefit is calculated over a data distribution with 30% positive cases at a decision threshold of 0.25 with different values for k.*

In practice there may be multiple decision thresholds of interest, in which case we can optimize the average σNB across these thresholds, or a weighted average if some thresholds are more important than others (11). In the following we only consider unweighted averages for threshold ranges and denote the σNB for threshold range $\mathcal{T}$ as $\sigma\text{NB}_{\mathcal{T},k}$.

## 2.2 Motivating 1D example: piece-wise linear function

To motivate when optimizing net benefit directly may be preferable to optimizing NLL, we consider a simple example where the true relationship between the feature $x$ and outcome is a piece-wise linear function in logit space with two connected linear parts, see Figure 2. If the data were modelled with standard logistic regression, the true function is not in the model class as logistic regression is a linear function in logit space. Depending on where the observations lie (around inflection point, far away from this point, uniform), logistic regression will fit different lines, which may yield suboptimal net benefit depending on the decision threshold. In contrast, directly optimizing net benefit around the relevant threshold would allow the model to match the corresponding line segment and achieve optimal net benefit. For $x$ uniformly distributed between -2 and 2 and inflection points in logit space of $\{(-2, -1.667), (0, -1), (2, 3)\}$, the consequence is an increase in NB of 0.036 for NLL and 0.063 for optimal NB, see Figure 2.

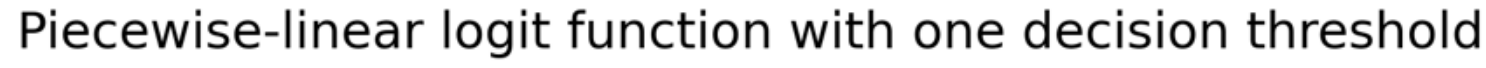

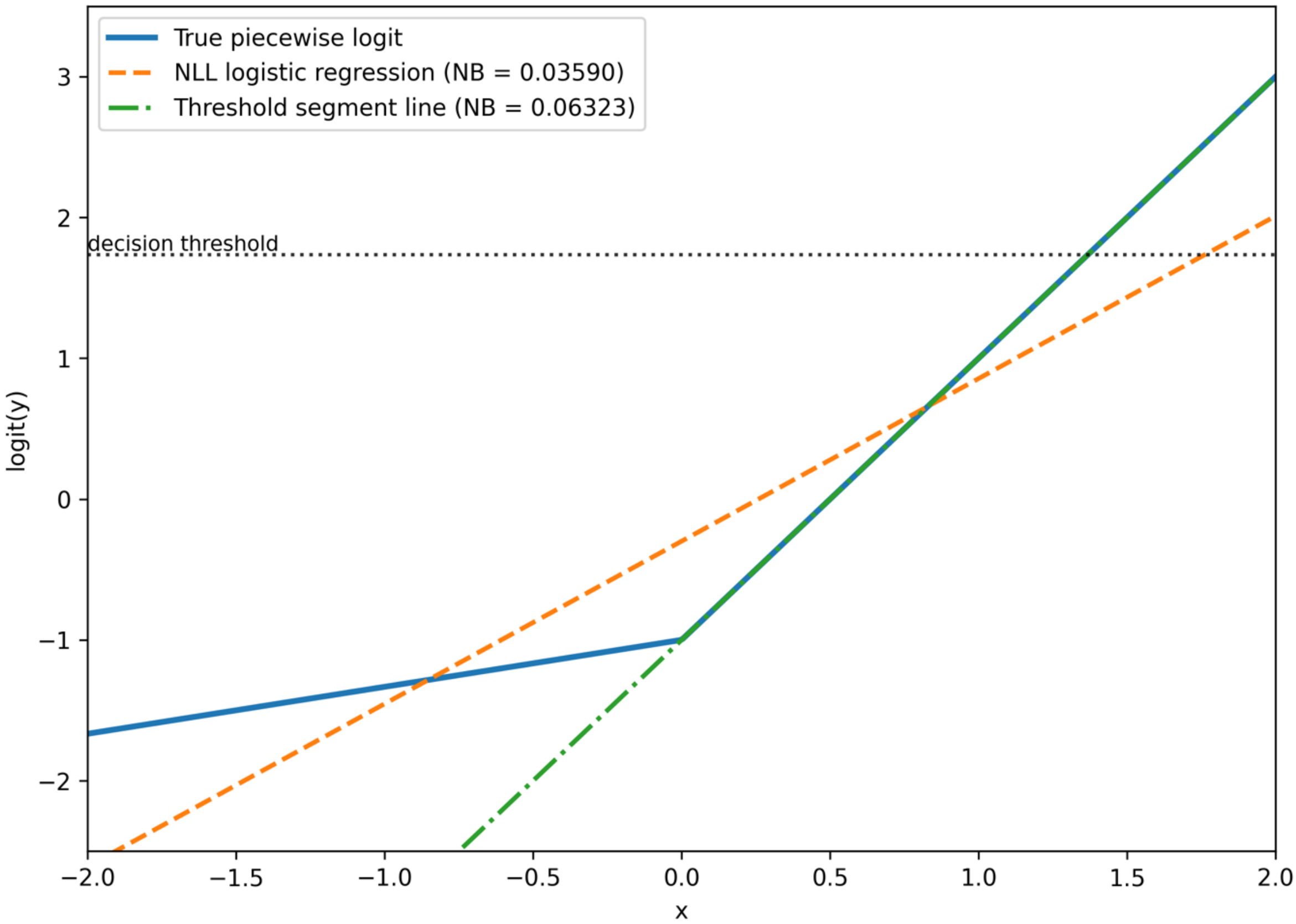

*Figure 2. Example of a data generating mechanism where the logit of the probability of the outcome is piecewise linear in feature X, with X uniformly distributed between -2 and 2. When fitting a standard logistic regression model (a single straight line), this will interpolate between the two line segments. Depending on where the decision threshold lies, better Net Benefit can be achieved by fitting either the first or the second line segment. Direct optimization of Net Benefit at the relevant threshold would target this. Net Benefit calculation achieved with numeric integration.*

## 2.3 Training procedure

After specifying the decision threshold $t$, or a range of relevant thresholds, we optimize σNB using gradient-based optimization procedures appropriate for each model class. Optimization of σNB requires balancing approximation bias (low inverse temperature $k$, high smoothness) against gradient informativeness. To manage this trade-off, we introduce a temperature annealing strategy in which $k$ is gradually increased during training according to the schedule $1 \rightarrow 4 \rightarrow 10$. This allows optimization to begin with a smooth objective that provides stable and informative gradients, while progressively transitioning toward a sharper approximation of net benefit.

In addition, the σNB objective is non-convex and may contain regions with near-zero gradients, particularly when model predictions lie far from the decision threshold. To address this, σNB optimization is initialized from a model pre-trained using NLL ('warm start'), providing a stable starting point in parameter space and reducing sensitivity to poor local optima.

Because continued optimization of the smooth surrogate does not necessarily improve thresholded NB, model checkpoints were selected using NB rather than the smooth training objective. For logistic regression and GAMs, checkpoints were selected using NB on the training data, whereas XGBoost used inner validation folds as described below. Final performance for all model classes was evaluated exclusively on held-out test folds that were not used for model fitting or checkpoint selection. The overall procedure combining NLL warm-start initialization, temperature annealing, and NB-based checkpoint selection is summarized in figure 3.

**Algorithm 1:** Smooth Net Benefit Training Algorithm

**Input:** $D_{\text{train}}, D_{\text{valid}}{}^{(*)},\ f(\cdot;\theta),\ \mathcal{T},\ \{k_j\}_{j=1}^{J},\ \varepsilon,\ patience$

**Output:** $\hat{\theta}$

1: Train $f(\cdot;\theta)$ using Bernoulli Negative Log Likelihood on $D_{\text{train}}$ to obtain warm-start parameters $\theta^{(0)}$.
2: Set $\hat{\theta} \leftarrow \theta^{(0)}$.
3: Initialize the best observed net benefit.
4: **for** $j = 1, \ldots, J$ **do**
5: $\quad$ $patience_counter \leftarrow 0$.
6: $\quad$ **while** $patience_counter < patience$ **do**
7: $\quad\quad$ Perform one $\sigma$NB training epoch on $D_{\text{train}}$ using threshold range $\mathcal{T}$ and inverse temperature $k_j$.
8: $\quad\quad$ Evaluate net benefit on $D_{\text{valid}}$ over threshold range $\mathcal{T}$.
9: $\quad\quad$ **if** *validation net benefit improves by more than* $\varepsilon$ **then**
10: $\quad\quad\quad$ Update the best observed net benefit.
11: $\quad\quad\quad$ Update $\hat{\theta}$.
12: $\quad\quad\quad$ $patience_counter \leftarrow 0$.
13: $\quad\quad$ **else**
14: $\quad\quad\quad$ $patience_counter \leftarrow patience_counter + 1$.
15: **return** $\hat{\theta}$

**Note:** $^{(*)}$ For logistic regression and GAM, $D_{valid} = D_{train}$; for XGBoost, $D_{valid}$ is a separate validation set.

*Figure 3. Smooth Net Benefit training algorithm. Overview of the training procedure combining NLL warm-start initialization, inverse-temperature annealing, and Net Benefit-based checkpoint selection*

### 2.3.1 Implementation details

We implement an σNB training procedure for three model classes: logistic regression, generalized additive models (GAMs) and XGBoost. For all three model classes, σNB optimization followed the same temperature annealing schedule, with the inverse temperature increasing from $k = 1$ to 4 and 10. The model-specific implementations are described below.

For logistic regression and GAMs, the NLL loss was replaced by the σNB objective while model structure and regularization were held fixed. NLL warm-start models were first fitted using L2 regularization parameters selected by five-fold cross-validation. The selected L2 penalty for logistic regression and the selected linear-term and spline smoothness penalties for GAMs were retained during σNB optimization, isolating the effect of the training objective from changes in regularization or model complexity. During σNB continuation, training-set NB was used for checkpoint selection.

For XGBoost, σNB was implemented as a custom objective. Because gradient-boosted trees rely on second-order optimization, three Hessian variants were considered: a positive semi-definite approximation based on absolute weights (“absw”), a constant positive approximation (“fixed025”), and the exact second derivative (“true”), which may be indefinite (12).

σNB optimization was applied as a continuation procedure from an NLL-trained XGBoost model within a nested cross-validation design. NLL hyperparameters were selected using inner-fold validation NLL, after which sNB continuation was warm-started from the corresponding NLL models. sNB-specific continuation hyperparameters and checkpoint length were selected using Net Benefit on the inner validation folds. The selected configuration was then refitted on the complete outer training data and evaluated exclusively on the held-out outer test fold. Detailed hyperparameter and checkpoint-selection procedures are provided in Supplementary Methods.

The implementation of Smooth Net Benefit and code used to reproduce the analyses are publicly available at the Smooth Net Benefit GitHub repository (https://github.com/KoenGorgels/Smooth-Net-Benefit).

#### 2.3.2 Post-hoc calibration comparators

As additional comparators we evaluated local Platt scaling as post-hoc decision-focused calibration baseline, and additionally a one-parameter simplified version of Platt scaling called ‘local temperature scaling’ (10,13). Both methods were fitted to predictions from the NLL-trained models using training observations around the target decision threshold and were subsequently applied to the corresponding test predictions. Performance was evaluated using Net Benefit over the same threshold band as for sNB. Details of local-subset construction and model fitting are provided in Supplementary Methods.

### 2.4 Experimental evaluation

#### 2.4.1 Framingham

The Framingham dataset was used to construct an illustrative dataset for predicting incident coronary heart disease (CHD) within 15 years. After excluding participants with prevalent CHD and those whose 15-year outcome could not be determined due to insufficient follow-up, the dataset comprised 3,781 individuals, of whom 630 (16.7%) experienced CHD within 15 years.

To assess the impact of σNB-based training, models were evaluated using one-hot encoding of categorical covariates. σNB training and evaluation were conducted over threshold ranges of ±0.025 centered at 5%, 10%, and 20% estimated 15-year cardiovascular disease risk. These thresholds were selected to span lower to higher clinically relevant cardiovascular risk levels, consistent with the risk-stratified approach used in European cardiovascular prevention guidelines (14).

Models were evaluated using five rotating 80/20 train-test splits. For each model and decision-threshold combination, Net Benefit was calculated on the held-out test fold and compared between methods using matched splits. Raw Net Benefit was used for the Framingham analyses. Mean paired differences in Net Benefit and their 95% confidence intervals were calculated across the five matched splits.

### 2.4.2 Benchmark: TabZilla

We evaluated σNB on a large tabular data benchmark suite (TabZilla) used in previous benchmarks (15). The TabZilla benchmark suite comprises a large collection of real-world and synthetic datasets spanning a wide range of sample sizes, and outcome prevalences. We restricted the analysis to binary classification datasets with at least 1,000 observations as well as datasets with fewer features than observations.

Each dataset was evaluated using five-fold stratified cross-validation. In each outer fold, four folds were used for model development and the remaining fold was held out for evaluation. Model selection and hyperparameter tuning were performed exclusively using the outer training data.

For each dataset, training and evaluation was done with the threshold set at the observed prevalence defined as the proportion of positive outcomes, and additionally at the inverse prevalence when the prevalence lay outside the 40% to 60% range, in order to study performance under both low- and high-threshold decision settings. Under the assumption that the bulk of observations will have underlying risks closer to the prevalence value instead of closer to 1 minus the prevalence, this mimics settings where either most observations are close to the decision threshold or far away. For each evaluation threshold, a range of width 0.05 (i.e., $\pm 0.025$) was constructed around the threshold, and net benefit was averaged over this range.

Preprocessing was fitted exclusively on the training portion of each outer split and subsequently applied to the corresponding held-out data. Binary variables were encoded directly, categorical variables were one-hot encoded after grouping uncommon levels, and continuous variables were median-imputed. Continuous predictors were standardized for logistic regression and XGBoost and represented using spline basis functions for GAMs. Full preprocessing rules are provided in Supplementary Methods.

Standardized Net Benefit, defined as Net Benefit divided by outcome prevalence, was the primary performance measure for the TabZilla benchmark. Corresponding raw Net Benefit results are provided in the Supplement.

## 2.5 Statistical analysis

For the Framingham analyses, performance was compared between methods using paired differences in Net Benefit across the five matched held-out splits. Mean paired differences and 95% confidence intervals were calculated from the split-level differences using the t distribution.

For the TabZilla benchmark, performance for each dataset–threshold combination was first averaged across the five held-out folds. Comparisons between training approaches were then performed within model families using paired differences in standardized Net Benefit across dataset–threshold combinations. Mean paired differences, 95% confidence intervals, and two-sided p-values were calculated using paired t-based inference, with the dataset–threshold combination as the unit of analysis. Raw Net Benefit was analyzed analogously and is reported in the Supplement. At the individual dataset–threshold level, 95% confidence intervals for paired differences were additionally calculated across the five held-out folds to characterize the direction and consistency of effects.

For the flexibility analysis, the association between flexibility gain and σNB gain in logistic regression was evaluated using linear regression adjusted for log-transformed sample size and number of features. Partial Pearson and Spearman correlations adjusted for the same covariates were also calculated. To reduce correlation induced by the shared NLL-trained logistic-regression comparator, flexibility gain and σNB gain were estimated from non-overlapping subsets of folds, with the allocation reversed in a complementary analysis.

All statistical tests were two-sided, and 95% confidence intervals excluding zero were considered evidence of a difference between methods.

# 3 Results

## 3.1 Illustrative dataset: Framingham

σNB training did not consistently improve Net Benefit relative to NLL across the Framingham experiments (Table 1). At the 5% threshold, the paired 95% confidence interval excluded zero for the small improvements observed after local Platt and local temperature scaling in logistic regression and GAMs; no other comparison showed clear evidence of improvement.

| **Model type** | **Model subtype** | **Threshold 5%** | **Threshold 10%** | **Threshold 20%** |
|---|---|---|---|---|
| **LR** | NLL | 0.1271 | 0.0967 | 0.0515 |
| | σNB | 0.1266 | 0.0948 | 0.0481 |
| | Platt scaling | 0.1274* | 0.0965 | 0.0514 |
| | Temp scaling | 0.1273* | 0.0964 | 0.0514 |
| **GAM** | NLL | 0.1271 | 0.0968 | 0.0514 |
| | σNB | 0.1266 | 0.0943 | 0.0493 |
| | Platt scaling | 0.1273* | 0.0966 | 0.0515 |
| | Temp scaling | 0.1274* | 0.0966 | 0.0514 |
| **XGBoost** | NLL | 0.1263 | 0.0952 | 0.0471 |
| | σNB (hessian type) | 0.1255 (absw)<br>0.1262 (fixed025)<br>0.1262 (true) | 0.0938 (absw)<br>0.0940(fixed025)<br>0.0933 (true) | 0.0442 (absw)<br>0.0442 (fixed025)<br>0.0470 (true) |
| | Platt scaling | 0.1255 | 0.0936 | 0.0462 |
| | Temp scaling | 0.1256 | 0.0931 | 0.0466 |

***Table 1.*** *Results for the Framingham dataset. LR = logistic regression; GAM = generalized additive model; XGBoost = extreme gradient boosting; NLL = negative log-likelihood; σNB = smooth Net Benefit; absw = absolute-weight Hessian approximation; fixed025 = constant Hessian approximation of 0.25; true = exact second derivative. An asterisk (*) indicates that the 95% confidence interval for the paired difference in Net Benefit relative to NLL training excluded zero.*

## 3.2 Benchmark: TabZilla

### Overall performance

A total of 44 datasets were included, yielding 72 unique dataset–threshold combinations when both prevalence and inverse-prevalence evaluation settings were considered. Dataset prevalence ranged from 5% to 94%, and sample size varied from 1,000 to 539,383 observations.

To investigate when σNB training is most beneficial relative to NLL training, we restricted the analysis to dataset–threshold combinations where differences between training objectives could be meaningfully interpreted. Specifically, we excluded combinations where both NLL- and σNB-trained logistic regression had negligible test net benefit (<0.0001) as these floor cases provide little scope for meaningful differences between the training objectives. We also excluded combinations where both methods achieved near-ceiling performance (>98% standardized test net benefit), because there was little room for further improvement. After applying these criteria, 59 dataset–threshold combinations remained for analysis. The results for all datasets without any exclusions are presented in the Appendix.

| Model type | NLL | σNB | Local temperature scaling | Local Platt scaling |
|---|---|---|---|---|
| **Logistic regression** | 0.5669 | 0.5765 | 0.5718 | 0.5689 |
| **GAM** | 0.5921 | 0.5625 | 0.5885 | 0.5862 |
| **XGBoost** | 0.6745 | 0.6723 (absw)<br>0.6727 (fixed025)<br>0.6735 (true) | 0.6535 | 0.6300 |

*Table 2. Mean standardized Net Benefit across the 59 dataset–threshold combinations in the restricted TabZilla analysis. NLL, negative log-likelihood; σNB, Smooth Net Benefit; GAM, generalized additive model; XGBoost, Extreme Gradient*

*Boosting; absw, absolute-weight Hessian approximation; fixed025, fixed Hessian approximation of 0.25; true, exact second derivative.*

Within model families, the effect of σNB training differed across model classes (Figure 4). For logistic regression, mean standardized Net Benefit increased from 0.5669 with NLL to 0.5765 with σNB (mean difference 0.0096, 95% CI −0.0001 to 0.0193). At the individual dataset–threshold level, fold-level 95% confidence intervals favored σNB in 19 combinations, NLL in 3, and included zero in 37. For GAMs, mean standardized Net Benefit decreased from 0.5921 to 0.5625 (mean difference −0.0296, 95% CI −0.0721 to 0.0129). None of the three σNB Hessian implementations improved standardized Net Benefit relative to NLL for XGBoost. Corresponding raw Net Benefit results are provided in the Supplement.

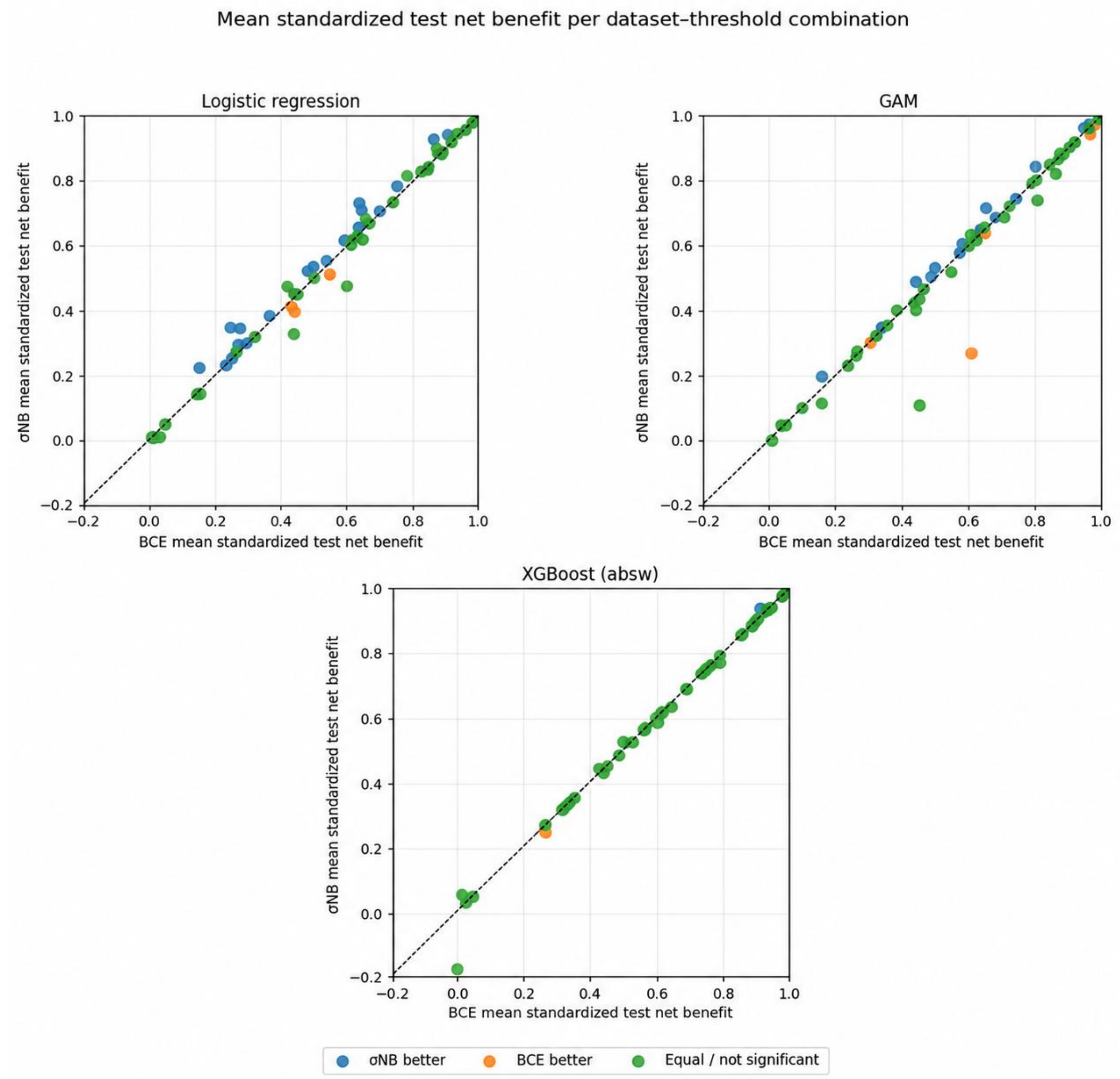

*Figure 4. Results on the TabZilla benchmark after excluding dataset–threshold combinations in which both NLL- and σNB-trained logistic regression models had either negligible test Net Benefit (<0.0001) or near-ceiling standardized test Net Benefit (>98%).*

Post-hoc local calibration did not consistently improve standardized Net Benefit relative to NLL training. For logistic regression, local temperature scaling produced a mean difference of 0.0049 (95% CI −0.0021 to 0.0119) and local Platt scaling a difference of 0.0021 (95% CI −0.0087 to 0.0129). Both methods slightly reduced mean performance for GAMs and produced larger reductions for XGBoost (Table 2).

## Flexibility and σNB gains

The motivating example suggested that σNB-based training may be most beneficial when model flexibility is limited. To investigate whether this pattern generalized across datasets, we examined whether the gain from σNB training in logistic regression was associated with the performance advantage of XGBoost over NLL-trained logistic regression, used as a proxy for the potential benefit of increased model flexibility. Because both quantities contain the performance of NLL-trained logistic regression, calculating them from the same test folds could induce a spurious positive association through shared sampling noise. The five folds were therefore divided into independent subsets: in one analysis, σNB gain was estimated using folds 1–3 and flexibility gain using folds 4–5, while the allocation was reversed in a complementary analysis. Only prevalence-threshold runs were included to ensure a consistent comparison across the included datasets. Of the 44 benchmark datasets, one was excluded because it met the near-ceiling performance criterion, leaving 43 datasets for the flexibility analysis.

After adjustment for log-transformed feature count and dataset size, flexibility gain was positively associated with σNB gain in both analyses using linear regression. Partial Pearson correlations were $r = 0.490 (p = 0.00084)$ and $r = 0.494 (p = 0.00077)$, while partial Spearman correlations were $\rho = 0.420 (p = 0.0050)$ and $\rho = 0.520 (p = 0.00035)$ (figure 5). Likewise, adding flexibility gain significantly improved the regression model beyond feature count and dataset size alone in both split A ($\chi^2(1) = 11.83, p = 0.00058$) and split B ($\chi^2(1) = 12.01, p = 0.00053$). Within these regression models, feature count was negatively associated with σNB gain, whereas dataset size showed no significant association.

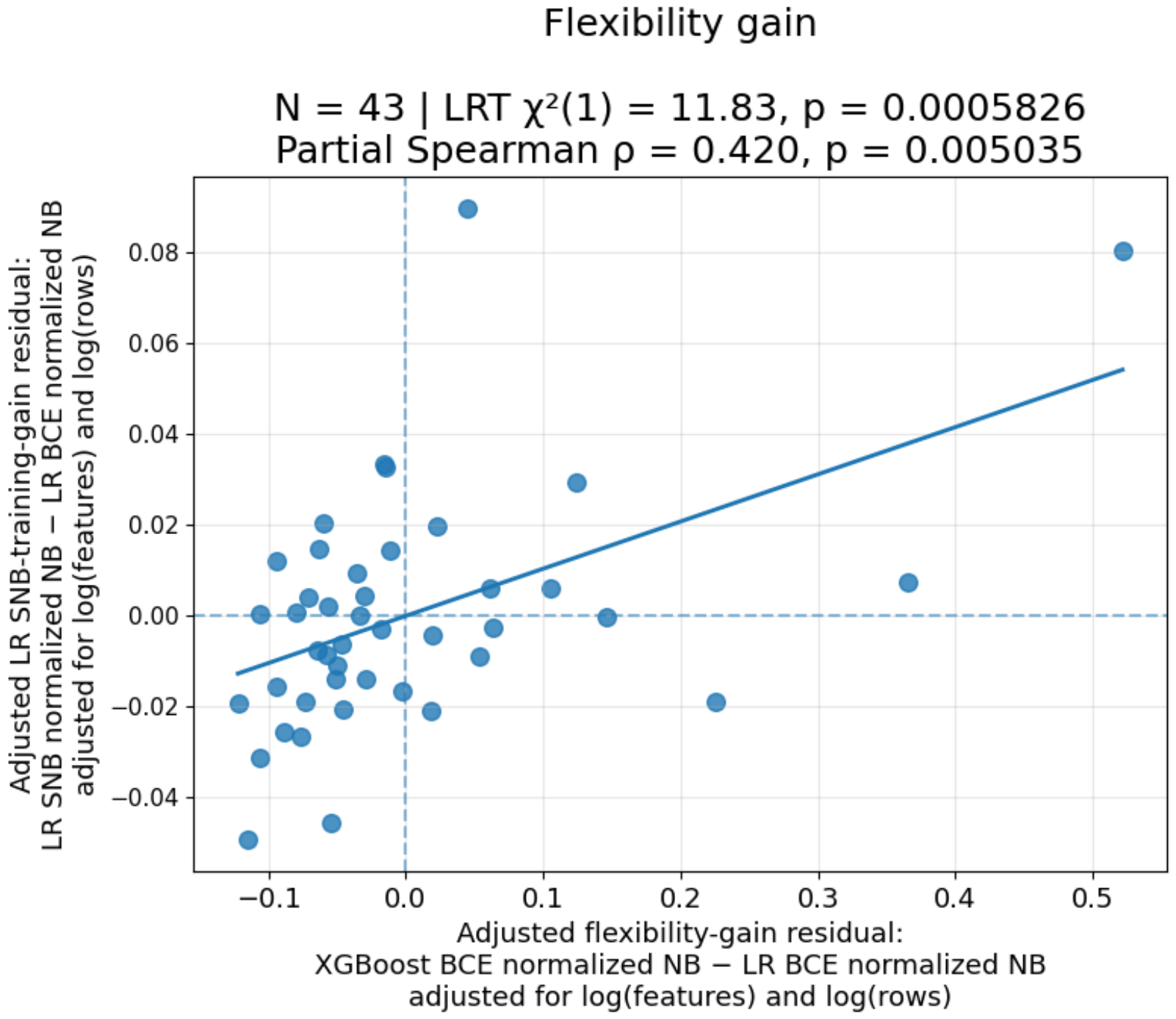

*Figure 5. Relationship between flexibility gain (XGBoost versus logistic regression) and the gain from σNB training in logistic regression across the 43 datasets evaluated at the prevalence threshold. Both quantities were estimated from non-overlapping subsets of folds to avoid shared sampling noise.*

### 3.2.1 Analysis on small datasets

Because the main benchmark contained several large datasets that may favor more flexible models, we conducted an additional small-sample analysis. Training sets were reduced to contain at least 200 positive and 200 negative observations while preserving the original outcome prevalence, with remaining observations used for testing; datasets unable to meet these criteria were excluded. This yielded 61 dataset–threshold combinations from 38 datasets and 50 combinations after applying the same near-floor and near-ceiling exclusion criteria as in the restricted main analysis. Mean standardized Net Benefit was 0.5542 for NLL-trained logistic regression and 0.5505 for σNB-trained logistic regression (mean difference −0.0037, 95% CI −0.0125 to 0.0050). XGBoost remained higher at 0.6006. Additional results and dataset sizes are provided in the Supplement.

# 4 Discussion

We introduced a smooth approximation of net benefit (σNB) as a training objective along with a training algorithm for logistic regression, GAMs and XGBoost, and evaluated its performance across multiple datasets. Across both the Framingham dataset and the TabZilla benchmark, σNB-based training did not yield consistent or substantial improvements over NLL, with only modest gains observed for logistic regression. In contrast, Net Benefit was primarily associated with model class, with XGBoost generally outperforming GAMs and logistic regression. These findings suggest that, in the settings considered, model flexibility may play a larger role in decision performance than directly aligning the training objective with the downstream decision metric.

The pattern observed for logistic regression suggests that the value of σNB may depend on model misspecification near the decision threshold. As illustrated by the motivating example, a linear model can provide a poor global approximation while still approximating a nonlinear relationship adequately within a local region. For threshold-based decisions, model behavior in the region where predicted risks cross the decision threshold is particularly consequential (9). σNB may therefore help when an NLL-trained logistic regression allocates its limited model capacity away from this decision-relevant region, while offering little additional benefit when the relevant structure is already captured. Consistent with this interpretation, σNB gains were positively associated with the performance advantage of XGBoost over NLL-trained logistic regression. Framingham showed a similar pattern: XGBoost did not outperform logistic regression substantially, and σNB likewise produced no consistent improvement.

The local calibration analyses provide additional context. If the modest logistic-regression gains with σNB primarily reflected better alignment of predicted risks around the decision threshold, decision-focused post-hoc calibration might be expected to yield similar improvements. Instead, local temperature and Platt scaling produced only small changes for logistic regression, no improvement for GAMs, and reduced Net Benefit for XGBoost. The σNB findings therefore do not appear to be explained by local recalibration alone and are consistent with the broader observation that additional decision-focused optimization offers limited benefit when NLL training already captures the relevant decision structure.

The results also illustrate a trade-off between objective alignment and optimization stability. Although σNB more directly reflects threshold-specific decision utility, it introduces a non-convex objective with limited gradient information away from the decision threshold and therefore requires warm-start initialization and temperature annealing. In contrast, NLL is a strictly proper scoring rule and, for logistic regression, yields a convex optimization problem (16). Thus, although NLL does not explicitly optimize threshold-specific decision utility, it provides stable probability estimation, which may explain its strong empirical performance when the model class is sufficiently flexible or well specified.

The modest logistic-regression advantage also disappeared in the small-sample analysis. Under reduced training-set sizes, σNB and NLL produced similar Net Benefit, suggesting that the potential benefit of decision-focused optimization is not robust to more limited data. In such settings, sampling variability and optimization instability may outweigh any advantage from more closely aligning the training objective with the downstream decision metric.

Several limitations remain. First, σNB constitutes a differentiable approximation to net benefit, and the associated trade-off between smoothness and fidelity may limit the extent to which optimization improvements translate into gains in thresholded decision performance. Second, in the benchmark analysis, evaluation focused on threshold ranges centered around the prevalence and its inverse, consistent with earlier work (17). Although the Framingham analyses did not suggest substantial sensitivity to alternative threshold choices, this restricted evaluation may limit generalizability to settings with substantially different or narrowly defined decision thresholds. Finally, no dedicated benchmark of large-scale medical datasets was included, reflecting the limited availability of sufficiently large public medical datasets suitable for prediction research.

For logistic regression and GAMs, regularization parameters selected during NLL model development were retained during σNB continuation to isolate the effect of changing the training objective. Consequently, these experiments evaluate the incremental effect of sNB within an NLL-tuned learner rather than a fully independently tuned sNB model. Objective-specific regularization could potentially improve sNB performance. However, logistic regression showed modest gains despite using NLL-selected regularization, whereas GAM did not, and XGBoost showed no improvement despite σNB-specific tuning. This pattern suggests that objective-specific tuning alone is unlikely to explain the main findings.

## Conclusion

Overall, σNB training did not consistently or substantially improve Net Benefit over standard NLL training. The clearest gains were observed for logistic regression, whereas GAMs and XGBoost showed little or no benefit. These findings do not support σNB as a general replacement for likelihood-based training but suggest that its value may depend on the interaction between model flexibility, model misspecification, optimization, and the decision threshold. Future work should focus on identifying settings in which direct optimization of decision-focused objectives provides meaningful advantages beyond those obtainable through increased model flexibility alone.

**Data Availability**

The code implementing Smooth Net Benefit and the code used to reproduce the analyses are publicly available in the Smooth Net Benefit GitHub repository: github.com/KoenGorgels/Smooth-Net-Benefit. The TabZilla datasets analyzed in this study are publicly available through their original repositories and OpenML. The Framingham data were obtained from the publicly available Framingham dataset described in the manuscript.

**Funding**

This research received no dedicated funding.

**Competing interests**

The authors declare no competing interests.

**Artificial intelligence use**

OpenAI ChatGPT was used during manuscript preparation to assist with language editing and the refinement of text and code. All scientific decisions, analyses, interpretation of results, and final manuscript content were reviewed and approved by the authors.

## Supplementary Methods

### S1. XGBoost model development and σNB continuation

XGBoost model development used nested cross-validation within each outer training split. For the NLL baseline, learning rate, minimum child weight, maximum depth, lambda, gamma, and number of trees were selected using five-fold inner cross-validation based on validation NLL. The outer test fold was not used during model development.

For each inner fold, σNB continuation was warm-started from the corresponding NLL-trained model. Additional trees were fitted using the σNB custom objective, and Net Benefit over the relevant threshold range was evaluated on the inner validation data at regular checkpoints. The checkpoint with the highest validation Net Benefit was retained; when continuation did not improve validation Net Benefit, the original NLL model was retained.

For the XGBoost implementation of the σNB custom objective, let $z_i$denote the raw model margin for observation $i$, such that

$$p_i = \sigma(z_i).$$

For threshold $t_j$, define the threshold weight as

$$w_{t_j} = \frac{t_j}{1 - t_j},$$

and the smooth intervention weight as

$$s_{ij} = \sigma\left(k[z_i - \text{logit}(t_j)]\right),$$

where $k$is the inverse-temperature parameter. The corresponding threshold-specific σNB contribution is

$$s_{ij}\left[y_i - (1 - y_i)w_{t_j}\right].$$

Because XGBoost minimizes its objective, the negative of this contribution was used as the loss. For one observation and one threshold, the gradient with respect to the raw model margin is

$$g_{ij} = -\left[y_i - (1 - y_i)w_{t_j}\right]ks_{ij}(1 - s_{ij}).$$

Three Hessian implementations were evaluated. The **true** variant used the exact second derivative of the negative σNB loss,

$$h_{ij}^{\text{true}} = -\left[y_i - (1 - y_i)w_{t_j}\right] k^2 s_{ij}(1 - s_{ij})(1 - 2s_{ij}),$$

which is not guaranteed to be positive. The **absw** variant instead used the non-negative curvature approximation

$$h_{ij}^{\text{absw}} = \mid y_i - (1 - y_i)w_{t_j} \mid k^2 s_{ij}(1 - s_{ij}),$$

while the **fixed025** variant used the constant positive Hessian

$$h_{ij}^{\text{fixed025}} = 0.25.$$

For optimization over a threshold range, gradient and Hessian contributions were averaged across all thresholds in the corresponding range before being supplied to XGBoost.

σNB-specific hyperparameters, including continuation-tree parameters and the number of additional trees at each inverse-temperature stage, were selected using the same inner cross-validation procedure. For the final outer-fold model, the NLL baseline was refitted on the complete outer training set using the selected hyperparameters and the median number of baseline trees across inner folds. σNB continuation was then performed using the selected configuration and the median number of additional trees selected at each inverse-temperature stage. Final performance was evaluated exclusively on the held-out outer test fold.

## S2. Local calibration comparators

For local Platt and temperature scaling, a calibration subset was defined around each target decision threshold using training-set predictions. The initial region extended ±10 percentage points around the threshold and was iteratively expanded until it contained at least 50 events and 50 non-events. If these criteria could not be satisfied locally, the full available probability range was used.

Local temperature scaling estimated a single scalar temperature parameter by minimizing NLL within this subset. Local Platt scaling fitted a logistic calibration model using the same subset. The fitted transformations were subsequently applied to the corresponding test predictions, and performance was evaluated using Net Benefit over the same threshold range used for the NLL and σNB comparisons.

## S3. TabZilla preprocessing

All preprocessing was estimated exclusively from the training portion of each outer split and subsequently applied to the corresponding held-out data. Features with exactly two unique values were treated as binary and encoded as a single indicator. String variables and numeric variables with at most 20 unique values were treated as categorical. For each categorical variable, the 10 most frequent training-set levels were retained and remaining levels were grouped into an “other” category before one-hot encoding.

Remaining numeric variables were treated as continuous and median-imputed using values estimated from the training data. Continuous predictors were standardized for logistic regression and XGBoost. For GAMs, continuous predictors were represented using spline basis functions.